\documentclass{article} 
\usepackage{nips15submit_e,times}
\usepackage{hyperref}
\usepackage{url}
\usepackage{bm, subfigure}
\usepackage{inputenc}
\usepackage{amsmath,amssymb,amsthm}
\usepackage{amsthm}
\usepackage{algorithm, algorithmic}
\usepackage{graphicx}
\usepackage[backgroundcolor=white,linecolor=red,bordercolor=white,textsize=tiny,textwidth=10mm]{todonotes}

\usepackage{pdfpages}

\newcommand{\ica}{\hspace{0.25cm}}

\title{Stochastic Expectation Propagation}

\author{
Yingzhen Li \\
University of Cambridge\\
Cambridge, CB2 1PZ, UK \\
\texttt{yl494@cam.ac.uk} \\
\And
Jos\'e Miguel Hern\'andez-Lobato\\
Harvard University \\
Cambridge, MA 02138 USA \\
\texttt{jmh@seas.harvard.edu}
\And
Richard E.~Turner \\
University of Cambridge\\
Cambridge, CB2 1PZ, UK \\
\texttt{ret26@cam.ac.uk} \\
}

\nipsfinalcopy 

\begin{document}

\maketitle

\begin{abstract}
Expectation propagation (EP) is a deterministic approximation algorithm that is often used to perform approximate Bayesian parameter learning. EP approximates the full intractable posterior distribution through a set of local approximations that are iteratively refined for each datapoint. EP can offer analytic and computational advantages over other approximations, such as Variational Inference (VI), and is the method of choice for a number of models. The local nature of EP appears to make it an ideal candidate for performing Bayesian learning on large models in large-scale dataset settings. However, EP has a crucial limitation in this context: the number of approximating factors needs to increase with the number of data-points, $N$, which often entails a prohibitively large memory overhead. This paper presents an extension to EP, called stochastic expectation propagation (SEP), that maintains a global posterior approximation (like VI) but updates it in a local way (like EP).  Experiments on a number of canonical learning problems using synthetic and real-world datasets indicate that SEP performs almost as well as full EP, but reduces the memory consumption by a factor of $N$. SEP is therefore ideally suited to performing approximate Bayesian learning in the large model, large dataset setting.

\end{abstract}

\section{Introduction}

Recently a number of methods have been developed for applying Bayesian learning to large datasets. Examples include sampling approximations \cite{ahn:distributedMCMC, bardenet:MCMC}, distributional approximations including stochastic variational inference (SVI) \cite{hoffman:svi} and assumed density filtering (ADF) \cite{miguel:pbp}, and approaches that mix distributional and sampling approximations \cite{gelman:dep,xu:sms}. 
One family of approximation method has garnered less attention in this regard: Expectation Propagation (EP) \cite{minka:ep, opper:ec}. EP constructs a posterior approximation by iterating simple local computations that refine factors which approximate the posterior contribution from each datapoint. At first sight, it therefore appears well suited to large-data problems: the locality of computation make the algorithm simple to parallelise and distribute, and good practical performance on a range of small data applications suggest that it will be accurate \cite{kuss:gpep,barthelme:ep_likelihood,cunningham:gaussianEP}. 
However the elegance of local computation has been bought at the price of prohibitive memory overhead that grows with the number of datapoints $N$, since local approximating factors need to be maintained for every datapoint, which typically incur the same memory overhead as the global approximation. The same pathology exists for the broader class of power EP (PEP) algorithms \cite{minka:powerep} that includes variational message passing \cite{winn:vmp}. In contrast, variational inference (VI) methods \cite{jordan:variational,beal:variational} utilise global approximations that are refined directly, which prevents memory overheads from scaling with $N$.

Is there ever a case for preferring EP (or PEP) to VI methods for large data?  We believe that there certainly is. First, EP can provide significantly more accurate approximations. It is well known that variational free-energy approaches are biased and often severely so \cite{turner+sahani:2011a} and for particular models the variational free-energy objective is pathologically ill-suited such as those with non-smooth likelihood functions \cite{cunningham:gaussianEP,turner+sahani:2011c}. Second, the fact that EP is truly local (to factors in the posterior distribution and not just likelihoods) means that it affords different opportunities for tractable algorithm design, as the updates can be simpler to approximate.

As EP appears to be the method of choice for some applications, researchers have attempted to push it to scale. One approach is to swallow the large computational burden and simply use large data structures to store the approximating factors (e.g.~TrueSkill \cite{herbrich:trueskill}). This approach can only be pushed so far. A second approach is to use ADF, a simple variant of EP that only requires a global approximation to be maintained in memory \cite{maybeck:adf}. ADF, however, provides poorly calibrated uncertainty estimates \cite{minka:ep} which was one of the main motivating reasons for developing EP in the first place. 
A third idea, complementary to the one described here, is to use approximating factors that have simpler structure (e.g.~low rank, \cite{qi+minka:sparseGP}). This reduces memory consumption (e.g.~for Gaussian factors from $\mathcal{O}(ND^2)$ to $\mathcal{O}(ND)$), but does not stop the scaling with $N$. Another idea uses EP to carve up the dataset \cite{gelman:dep,xu:sms} using approximating factors for collections of datapoints. This results in coarse-grained, rather than local, updates and other methods must be used to compute them. (Indeed, the spirit of \cite{gelman:dep,xu:sms} is to extend sampling methods to large datasets, not EP itself.) 

Can we have the best of both worlds? That is, accurate global approximations that are derived from truly local computation. To address this question we develop an algorithm based upon the standard EP and ADF algorithms that maintains a global approximation which is updated in a local way.
%
We call this class of algorithms Stochastic Expectation Propagation (SEP) since it updates the global approximation with (damped) stochastic estimates on data sub-samples in an analogous way to SVI. Indeed, the generalisation of the algorithm to the PEP setting directly relates to SVI. 
Importantly, SEP reduces the memory footprint by a factor of $N$ when compared to EP. We further extend the method to control the granularity of the approximation, and to treat models with latent variables without compromising on accuracy or unnecessary memory demands. Finally, we demonstrate the scalability and accuracy of the method on a number of real world and synthetic datasets.



\section{Expectation Propagation and Assumed Density Filtering}
We begin by briefly reviewing the EP and ADF algorithms upon which our new method is based. Consider for simplicity observing a dataset comprising $N$ i.i.d.~samples $\mathcal{D} = \{\bm{x}_n \}_{n=1}^N$ from a probabilistic model $p(\bm{x}|\bm{\theta})$ parametrised by an unknown $D$-dimensional vector $\bm{\theta}$ that is drawn from a prior $p_0(\bm{\theta})$. Exact Bayesian inference involves computing the (typically intractable) posterior distribution of the parameters given the data, 
\begin{equation}
p(\bm{\theta} | \mathcal{D}) \propto p_0(\bm{\theta}) \prod_{n=1}^{N} p(\bm{x}_n | \bm{\theta}) \approx q(\bm{\theta}) \propto p_0(\bm{\theta}) \prod_{n=1}^{N} f_n(\bm{\theta}).
\end{equation}
Here $q(\bm{\theta})$  is a simpler tractable approximating distribution that will be refined by EP.
The goal of EP is to refine the approximate factors so that they capture the contribution of each of the likelihood terms to the posterior i.e.~$f_n(\bm{\theta}) \approx p(\bm{x}_n | \bm{\theta})$. In this spirit, one approach would be to find each approximating factor $f_n(\bm{\theta})$ by minimising the Kullback-Leibler (KL) divergence between the posterior and the distribution formed by replacing one of the likelihoods by its corresponding approximating factor,  $\mathrm{KL}[p(\bm{\theta}|\mathcal{D}) || p(\bm{\theta}|\mathcal{D}) f_n(\bm{\theta})/ p(\bm{x}_n | \bm{\theta})]$. Unfortunately, such an update is still intractable as it involves computing the full posterior. Instead, EP approximates this procedure by replacing the exact leave-one-out posterior $p_{-n}(\bm{\theta}) \propto p(\bm{\theta}|\mathcal{D}) / p(\bm{x}_n | \bm{\theta})$ on both sides of the KL by the approximate leave-one-out posterior (called the cavity distribution) $q_{-n}(\bm{\theta}) \propto q(\bm{\theta})/f_n(\bm{\theta})$. Since this couples the updates for the approximating factors, the updates must now be iterated.

In more detail, EP iterates four simple steps. First, the factor selected for update is removed from the approximation to produce the cavity distribution. Second, the corresponding likelihood is included to produce the tilted distribution $\tilde{p}_n(\bm{\theta}) \propto q_{-n}(\bm{\theta}) p(\bm{x}_n | \bm{\theta})$. Third EP updates the approximating factor by minimising $\mathrm{KL}[\tilde{p}_n(\bm{\theta}) || q_{-n}(\bm{\theta})  f_n(\bm{\theta})]$. The hope is that the contribution the true-likelihood makes to the posterior is similar to the effect the same likelihood has on the tilted distribution. If the approximating distribution is in the exponential family, as is often the case, then the KL minimisation reduces to a moment matching step \cite{amari:ig} that we denote $f_n(\bm{\theta}) \leftarrow \mathtt{proj}[\tilde{p}_n(\bm{\theta})] / q_{-n}(\bm{\theta}) $. Finally, having updated the factor, it is included into the approximating distribution.

We summarise the update procedure for a single factor in Algorithm \ref{alg:ep}. Critically, the approximation step of EP involves local computations since one likelihood term is treated at a time. The assumption is that these local computations, although possibly requiring further approximation, are far simpler to handle compared to the full posterior $p(\bm{\theta}| \mathcal{D})$. In practice, EP often performs well when the updates are parallelised. Moreover, by using approximating factors for groups of datapoints, and then running additional approximate inference algorithms to perform the EP updates (which could include nesting EP), EP carves up the data making it suitable for distributed approximate inference.

\begin{figure}[!t]
\begin{minipage}[t]{0.33\linewidth}
\centering
\begin{algorithm}[H] 
\caption{EP} \small
\label{alg:ep} 
\begin{algorithmic}[1] 
	\STATE choose a factor $f_n$ to refine:
	\STATE compute cavity distribution \\$q_{-n}(\bm{\theta}) \propto q(\bm{\theta}) / f_n(\bm{\theta})$ 
	\STATE compute tilted distribution \\$\tilde{p}_n(\bm{\theta}) \propto p(\bm{x}_n|\bm{\theta}) q_{-n}(\bm{\theta})$
	\STATE moment matching: \\ \hspace{-1mm}$f_n(\bm{\theta}) \leftarrow \mathtt{proj}[\tilde{p}_n(\bm{\theta})] / q_{-n}(\bm{\theta}) $
	\STATE inclusion:\\ $q(\bm{\theta}) \leftarrow q_{-n}(\bm{\theta}) f_n(\bm{\theta})$\\\hspace{1mm}\\ \vspace{1.5mm} \hspace{1mm}\\
\end{algorithmic}
\end{algorithm}
\end{minipage}
\begin{minipage}[t]{0.33\linewidth}
\centering
\begin{algorithm}[H] 
\caption{ADF} \small
\label{alg:adf} 
\begin{algorithmic}[1] 
	\STATE choose a datapoint $\bm{x}_n\sim \mathcal{D}$:
	\STATE compute cavity distribution \\$q_{-n}(\bm{\theta}) = q(\bm{\theta})$
	\STATE compute tilted distribution \\$\tilde{p}_n(\bm{\theta}) \propto p(\bm{x}_n|\bm{\theta}) q_{-n}(\bm{\theta})$
	\STATE moment matching: \\ \hspace{-1mm}$f_n(\bm{\theta}) \leftarrow \mathtt{proj}[\tilde{p}_n(\bm{\theta})] / q_{-n}(\bm{\theta}) $
	\STATE inclusion:\\ $q(\bm{\theta}) \leftarrow q_{-n}(\bm{\theta}) f_n(\bm{\theta})$\\\hspace{1mm}\\ \vspace{1.5mm} \hspace{1mm}\\
\end{algorithmic}
\end{algorithm}
\end{minipage}
\begin{minipage}[t]{0.33\linewidth}
\centering
\begin{algorithm}[H]
\caption{SEP} \small
\label{alg:sep} 
\begin{algorithmic}[1] 
	\STATE choose a datapoint $\bm{x}_n\sim \mathcal{D}$:
	\STATE compute cavity distribution \\ $q_{-1}(\bm{\theta}) \propto q(\bm{\theta}) / f(\bm{\theta})$
	\STATE compute tilted distribution \\$\tilde{p}_n(\bm{\theta}) \propto p(\bm{x}_n|\bm{\theta}) q_{-1}(\bm{\theta})$
	\STATE moment matching: \\\hspace{-1mm}$f_n(\bm{\theta}) \leftarrow \mathtt{proj}[\tilde{p}_n(\bm{\theta})] / q_{-1}(\bm{\theta}) $
	\STATE inclusion:\\ $q(\bm{\theta}) \leftarrow q_{-1}(\bm{\theta}) f_n(\bm{\theta})$
	\STATE \textit{implicit update}:\\ $f(\bm{\theta}) \leftarrow f(\bm{\theta})^{1 - \frac{1}{N}} f_n(\bm{\theta})^{\frac{1}{N}}$
\end{algorithmic}
\end{algorithm}
\end{minipage} 
\caption{Comparing the Expectation Propagation (EP), Assumed Density Filtering (ADF), and Stochastic Expectation Propagation (SEP) update steps. Typically, the algorithms will be initialised using $q(\bm{\theta}) = p_0(\bm{\theta})$ and, where appropriate, $f_n(\bm{\theta})=1$ or $f(\bm{\theta})=1$.}
\end{figure}

There is, however, one wrinkle that complicates deployment of EP at scale. Computation of the cavity distribution requires removal of the current approximating factor, which means any implementation of EP must store them explicitly necessitating an $\mathcal{O}(N)$ memory footprint. One option is to simply ignore the removal step replacing the cavity distribution with the full approximation, resulting in the ADF algorithm (Algorithm \ref{alg:adf}) that needs only maintain a global approximation in memory. But as the moment matching step now over-counts the underlying approximating factor (consider the new form of the objective $\mathrm{KL}[q(\bm{\theta}) p(\bm{x}_n | \bm{\theta}) || q(\bm{\theta})]$) the variance of the approximation shrinks to zero as multiple passes are made through the dataset. Early stopping is therefore required to prevent overfitting and generally speaking ADF does not return uncertainties that are well-calibrated to the posterior. 
In the next section we introduce a new algorithm that sidesteps EP's large memory demands whilst avoiding the pathological behaviour of ADF. 

\section{Stochastic Expectation Propagation}
%
In this section we introduce a new algorithm, inspired by EP, called Stochastic Expectation Propagation (SEP) that combines the benefits of local approximation (tractability of updates, distributability, and parallelisability) with global approximation (reduced memory demands).  The algorithm can be interpreted as a version of EP in which the approximating factors are tied, or alternatively as a corrected version of ADF that prevents overfitting. 
The key idea is that, at convergence, the approximating factors in EP can be interpreted as parameterising a global factor,  $f(\bm{\theta})$, that captures the average effect of a likelihood on the posterior  $f(\bm{\theta})^{N} \buildrel\triangle\over = \prod_{n=1}^{N} f_n(\bm{\theta}) \approx \prod_{n=1}^{N} p(\bm{x}_n | \bm{\theta})$. In this spirit, the new algorithm employs direct iterative refinement of a global approximation comprising the prior and $N$ copies of a single approximating factor, $f(\bm{\theta})$, that is $q(\bm{\theta}) \propto f(\bm{\theta})^N p_0(\bm{\theta})$.

SEP uses updates that are analogous to EP's in order to refine $f(\bm{\theta})$ in such a way that it captures the average effect a likelihood function has on the posterior. First the cavity distribution is formed by removing one of the copies of the factor, $q_{-1}(\bm{\theta}) \propto q(\bm{\theta})/f(\bm{\theta})$. 
Second, the corresponding likelihood is included to produce the tilted distribution $\tilde{p}_n(\bm{\theta}) \propto q_{-1}(\bm{\theta}) p(\bm{x}_n | \bm{\theta})$ and, third, SEP finds an intermediate factor approximation by moment matching, $f_n(\bm{\theta}) \leftarrow \mathtt{proj}[\tilde{p}_n(\bm{\theta})] / q_{-1}(\bm{\theta}) $. Finally, having updated the factor, it is included into the approximating distribution. It is important here not to make a full update since $f_n(\bm{\theta})$ captures the effect of just a single likelihood function  $p(\bm{x}_n | \bm{\theta})$. Instead, damping should be employed to make a partial update $f(\bm{\theta}) \leftarrow f(\bm{\theta})^{1 - \epsilon} f_n(\bm{\theta})^{\epsilon}$. A natural choice uses $\epsilon = 1/N$ which can be interpreted as minimising  $\mathrm{KL}[\tilde{p}_n(\bm{\theta}) || p_{0}(\bm{\theta})  f(\bm{\theta})^N]$ in the moment update, but other choices of $\epsilon$ may be more appropriate, including decreasing $\epsilon$ according to the Robbins-Monro condition \cite{robbins:stochastic}.

SEP is summarised in Algorithm \ref{alg:sep}. Unlike ADF, the cavity is formed by dividing out $f(\bm{\theta})$ which captures the average affect of the likelihood and prevents the posterior from collapsing. Like ADF, however, SEP only maintains the global approximation $q(\bm{\theta})$ since $f(\bm{\theta}) \propto (q(\bm{\theta}) / p_0(\bm{\theta}))^{\frac{1}{N}}$ and $q_{-1}(\bm{\theta}) \propto q(\bm{\theta})^{1 - \frac{1}{N}} p_0(\bm{\theta})^{\frac{1}{N}}$. When Gaussian approximating factors are used, for example, SEP reduces the storage requirement of EP from  $\mathcal{O}(ND^2)$ to $\mathcal{O}(D^2)$ which is a substantial saving that enables models with many parameters to be applied to large datasets.

\section{Algorithmic extensions to SEP and theoretical results}
SEP has been motivated from a practical perspective by the limitations inherent in EP and ADF. In this section we extend SEP in four orthogonal directions relate SEP to SVI. Many of the algorithms described here are summarised in Figure \ref{fig:relationship-algorithms} and they are detailed in the supplementary material.

\subsection{Parallel SEP: relating the EP fixed points to SEP}

The SEP algorithm outlined above approximates one likelihood at a time which can be computationally slow. However, it is simple to parallelise the SEP updates by following the same recipe by which EP is parallelised. Consider a minibatch comprising $M$ datapoints (for a full parallel batch update use $M=N$). First we form the cavity distribution for each likelihood. Unlike EP these are all identical. Next, in parallel, compute $M$ intermediate factors $f_m(\bm{\theta}) \leftarrow \mathtt{proj}[\tilde{p}_m(\bm{\theta})] / q_{-1}(\bm{\theta})$. In EP these intermediate factors become the new likelihood approximations and the approximation is updated to $q(\bm{\theta}) = p_0(\bm{\theta}) \prod_{n \ne m} f_n(\bm{\theta}) \prod_{m} f_m(\bm{\theta}) $. In SEP, the same update is used for the approximating distribution, which becomes $q(\bm{\theta}) \leftarrow p_0(\bm{\theta}) f_{old}(\bm{\theta})^{N-M} \prod_{m} f_m(\bm{\theta}) $ and, by implication, the approximating factor is $f_{new}(\bm{\theta}) = f_{old}(\bm{\theta})^{1-M/N} \prod_{m=1}^M f_m(\bm{\theta})^{1/N}$. One way of understanding parallel SEP is as a double loop algorithm. The \textbf{inner loop} produces intermediate approximations  $q_m(\bm{\theta}) \leftarrow \arg\min_q \mathrm{KL}[\tilde{p}_m(\bm{\theta}) ||q(\bm{\theta})]$; these are then combined in the \textbf{outer loop}: $q(\bm{\theta}) \leftarrow \arg\min_q \sum_{m=1}^M \mathrm{KL}[q(\bm{\theta}) ||q_m(\bm{\theta})] + (N-M) \mathrm{KL}[q(\bm{\theta}) || q_{old}(\bm{\theta})]$.

For $M=1$ parallel SEP reduces to the original SEP algorithm. For $M=N$ parallel SEP is equivalent to the so-called Averaged EP algorithm proposed in \cite{barthelme:aep} as a theoretical tool to study the convergence properties of normal EP. This work showed that, under fairly restrictive conditions (likelihood functions that are log-concave and varying slowly as a function of the parameters), AEP converges to the same fixed points as EP in the large data limit ($N \rightarrow \infty$).

There is another illuminating connection between SEP and AEP. Since SEP's approximating factor $f(\bm{\theta})$ converges to the geometric average of the intermediate factors $\bar{f}(\bm{\theta}) \propto [\prod_{n=1}^N f_n(\bm{\theta})]^{\frac{1}{N}}$, SEP converges to the same fixed points as AEP if the learning rates satisfy the Robbins-Monro condition \cite{robbins:stochastic}, and therefore under certain conditions \cite{barthelme:aep}, to the same fixed points as EP.
But it is still an open question whether there are more direct relationships between EP and SEP.

\subsection{Stochastic power EP: relationships to variational methods}
%
The relationship between variational inference and stochastic variational inference \cite{hoffman:svi} mirrors the relationship between EP and SEP. 
Can these relationships be made more formal? If the moment projection step in EP is replaced by a natural parameter matching step then the resulting algorithm is equivalent to the Variational Message Passing (VMP) algorithm \cite{minka:divergence} (and see supplementary material). Moreover, VMP has the same fixed points as variational inference \cite{winn:vmp} (since minimising the local variational KL divergences is equivalent to minimising the global variational KL). 

These results carry over to the new algorithms with minor modifications. Specifically VMP can be transformed into SVMP by replacing VMP's local approximations with the global form employed by SEP. In the supplementary material we show that this algorithm is an instance of standard SVI and that it therefore has the same fixed points as VI when $\epsilon$ satisfies the Robbins-Monro condition \cite{robbins:stochastic}. 
More generally, the procedure can be applied any member of the power EP (PEP) \cite{minka:powerep} family of algorithms which replace the moment projection step in EP with alpha-divergence minimization \cite{amari:ig}, but care has to be taken when taking the limiting cases (see supplementary).
These results lend weight to the view that SEP is a natural stochastic generalisation of EP.

\begin{figure}
\centering
\def\svgwidth{1\linewidth}
\includegraphics[height=6.5cm]{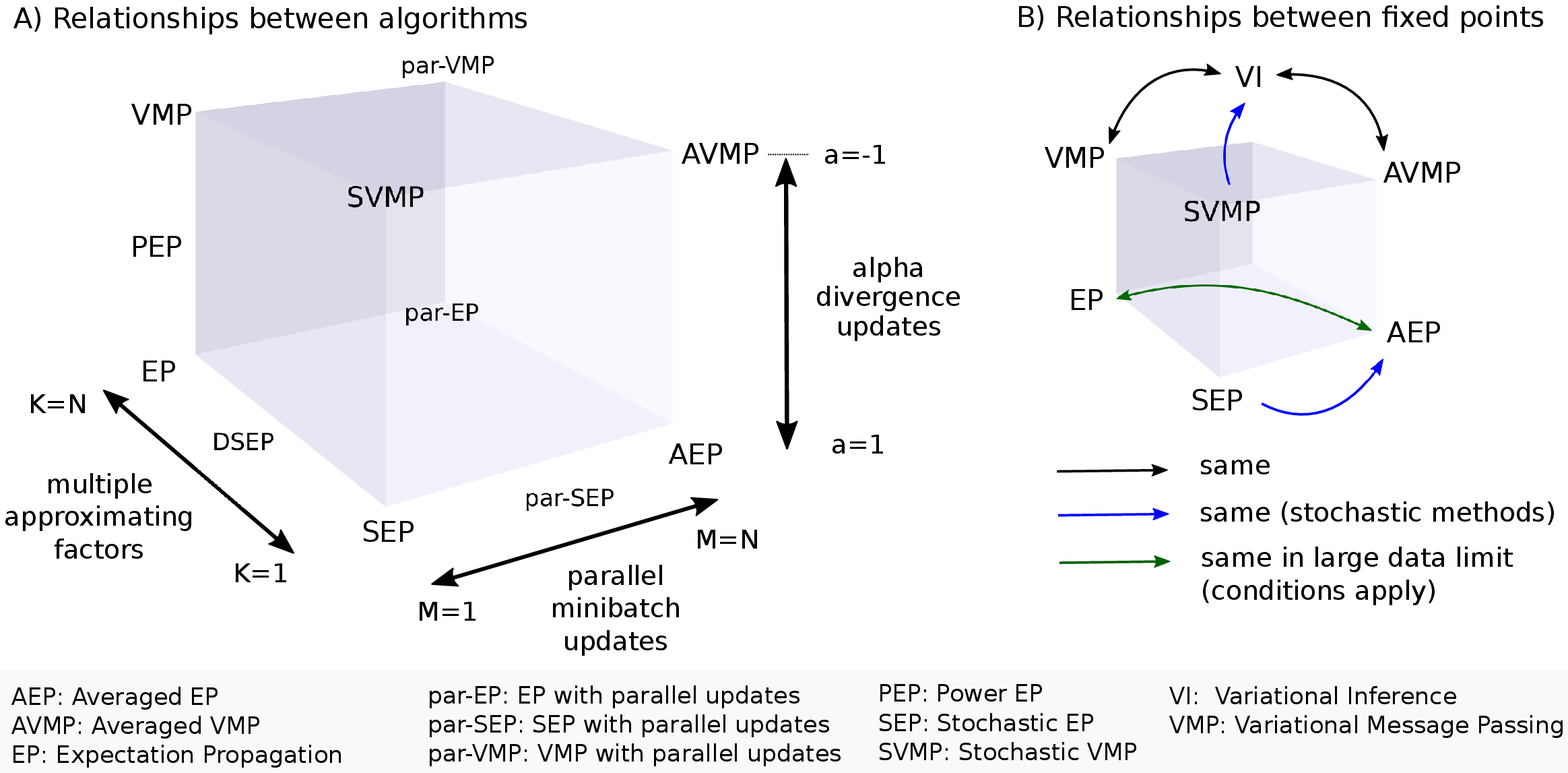}
\caption{Relationships between algorithms. Note that care needs to be taken when interpreting the alpha-divergence as $a \rightarrow -1$ (see supplementary material).}
\label{fig:relationship-algorithms}
\end{figure}

\subsection{Distributed SEP: controlling granularity of the approximation}

EP uses a fine-grained approximation comprising a single factor for each likelihood. SEP, on the other hand, uses a coarse-grained approximation comprising a signal global factor to approximate the average effect of all likelihood terms. One might worry that SEP's approximation is too severe if the dataset contains sets of datapoints that have very different likelihood contributions (e.g.~for odd-vs-even handwritten digits classification consider the affect of a 5 and a 9 on the posterior). It might be more sensible in such cases to partition the dataset into $K$ disjoint pieces $\{ \mathcal{D}_k = \{\bm{x}_n\}_{n=N_{k-1}}^{N_k} \}_{k=1}^{K}$ with $N = \sum_{k=1}^K N_k$ and use an approximating factor for each partition. If normal EP updates are performed \textbf{on} the subsets, i.e.~treating $p(\mathcal{D}_k|\bm{\theta})$ as a single true factor to be approximated, we arrive at the Distributed EP algorithm \cite{gelman:dep, xu:sms}. But such updates are challenging as multiple likelihood terms must be included during each update necessitating additional approximations (e.g.~MCMC). A simpler alternative uses SEP/AEP \textbf{inside} each partition, implying a posterior approximation of the form $q(\bm{\theta}) \propto p_0(\bm{\theta}) \prod_{k=1}^K f_{k}(\bm{\theta})^{N_k}$ with $f_{k}(\bm{\theta})^{N_k}$ approximating $p(\mathcal{D}_k|\bm{\theta})$. The limiting cases of this algorithm, when $K=1$ and $K=N$, recover SEP and EP respectively.

\subsection{SEP with latent variables}

Many applications of EP involve latent variable models. Although this is not the main focus of the paper, we show that SEP is applicable in this case without scaling the memory footprint with $N$. 
Consider a model containing hidden variables, $\bm{h}_n$, associated with each observation $p(\bm{x}_n, \bm{h}_n | \bm{\theta})$  that are drawn i.i.d.~from a prior $p_0(\bm{h}_n)$. The goal is to approximate the true posterior over parameters and hidden variables $p(\bm{\theta}, \{ \bm{h}_n\} | \mathcal{D}) \propto p_0(\bm{\theta}) \prod_n p_0(\bm{h}_n) p(\bm{x}_n | \bm{h}_n, \bm{\theta})$. 
Typically, EP would approximate the effect of each intractable term as $p(\bm{x}_n | \bm{h}_n, \bm{\theta})p_0(\bm{h}_n)  \approx f_n(\bm{\theta}) g_n(\bm{h}_n)$. Instead, SEP ties the approximate parameter factors $p(\bm{x}_n | \bm{h}_n, \bm{\theta})p_0(\bm{h}_n)  \approx f(\bm{\theta}) g_n(\bm{h}_n) $ yielding:
\begin{equation}
q(\bm{\theta}, \{ \bm{h}_n\}) \buildrel\triangle\over \propto p_0(\bm{\theta}) f(\bm{\theta})^N \prod_{n=1}^N g_n(\bm{h}_n) .
\end{equation}
Critically, as proved in supplementary, the local factors $g_n(\bm{h}_n)$ do not need to be maintained in memory. This means that all of the advantages of SEP carry over to more complex models involving latent variables, although this can potentially increase computation time in cases where updates for $g_n(\bm{h}_n)$ are not analytic, since then they will be initialised from scratch at each update.



\section{Experiments}

The purpose of the experiments was to evaluate SEP on a number of datasets (synthetic and real-world, small and large) and on a number of models (probit regression, mixture of Gaussians and Bayesian neural networks).

\subsection{Bayesian probit regression}
\begin{figure}
\centering
\def\svgwidth{0.31\linewidth}
\subfigure[\label{fig:sep_probit}]{
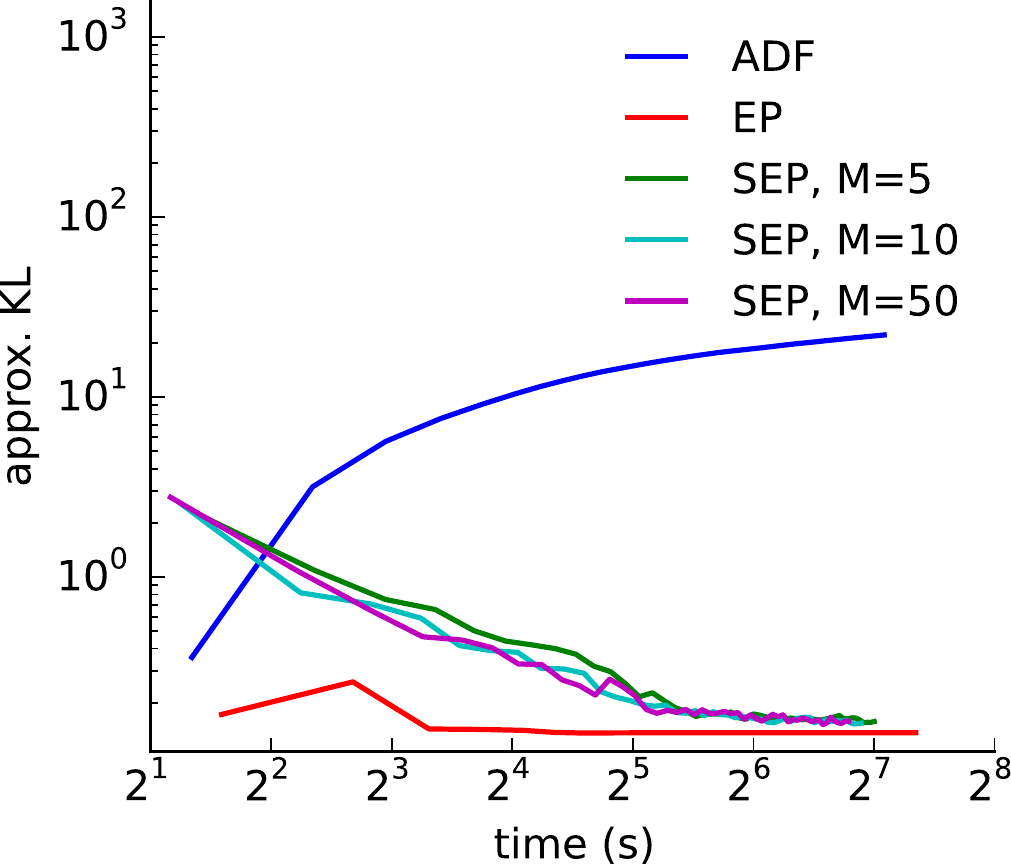}
%
%
\def\svgwidth{0.31\linewidth}
\subfigure[\label{fig:daep_probit}]{
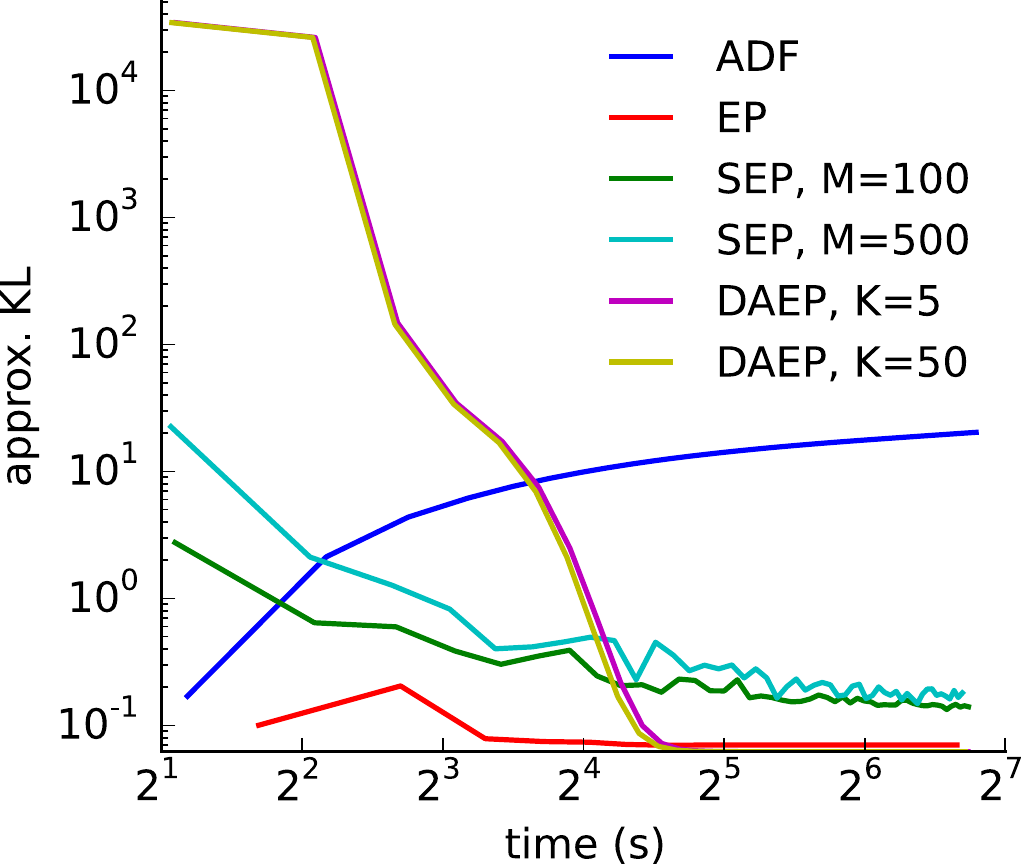}
%
%
\def\svgwidth{0.31\linewidth}
\subfigure[\label{fig:mnist}]{
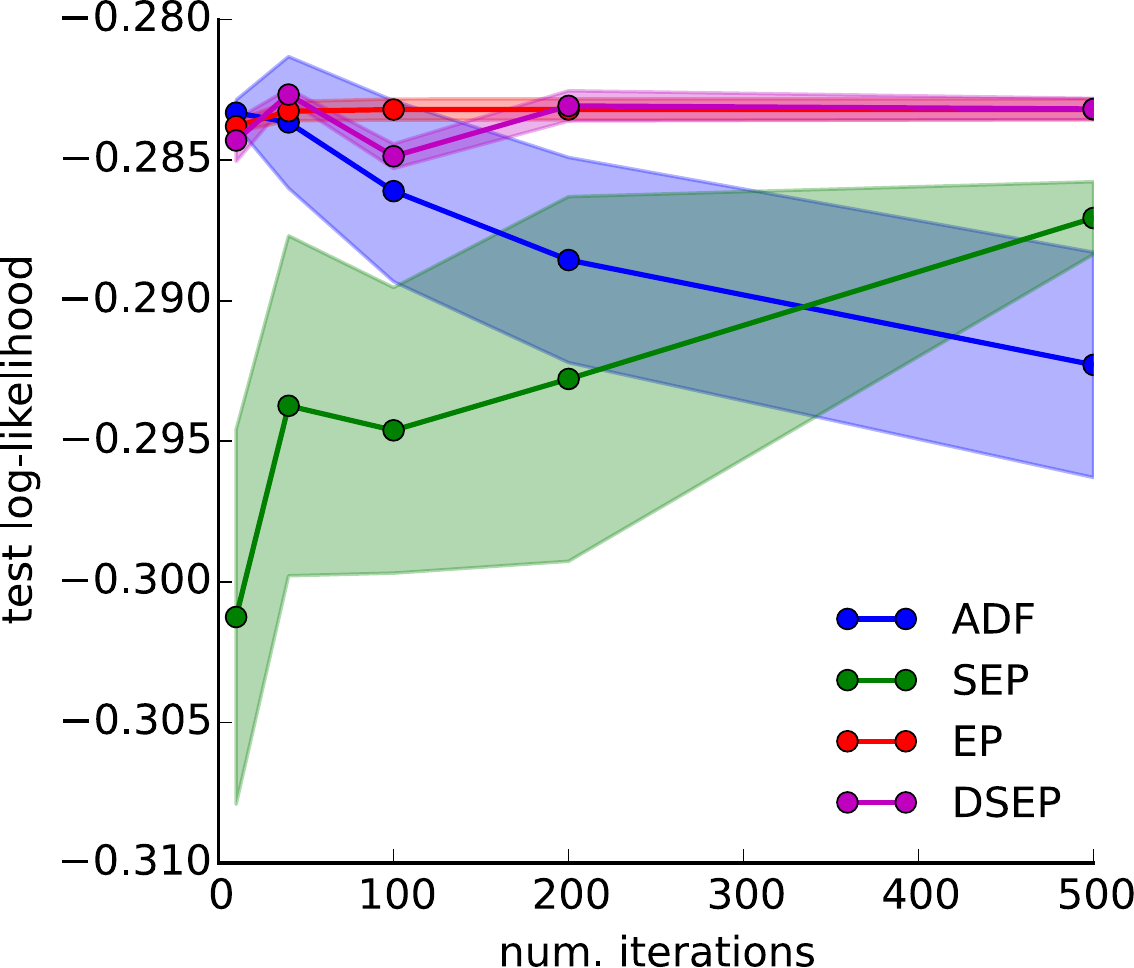}
\caption{Bayesian logistic regression experiments. Panels (a) and (b) show synthetic data experiments. Panel (c) shows the results on MNIST (see text for full details).}
\end{figure}

The first experiments considered a simple Bayesian classification problem and investigated the stability and quality of SEP in relation to EP and ADF as well as the effect of using mini-batches and varying the granularity of the approximation. The model comprised a probit likelihood function $P(\bm{y}_n = 1|\theta) = \Phi(\bm{\theta}^T \bm{x}_n)$ and a Gaussian prior over the hyper-plane parameter  $p(\bm{\theta}) = \mathcal{N}(\bm{\theta}; \bm{0}, \gamma I)$.  
The synthetic data comprised $N=5,000$ datapoints $\{ (\bm{x}_n, \bm{y}_n) \}$, where $\bm{x}_n$ were $D=4$ dimensional and were either sampled from a single Gaussian distribution (Fig.~\ref{fig:sep_probit}) or from a mixture of Gaussians (MoGs) with $J=5$ components (Fig.~\ref{fig:daep_probit}) to investigate the sensitivity of the methods to the homogeneity of the dataset. The labels were produced by sampling from the generative model. We followed \cite{xu:sms} measuring the performance by computing an approximation of $\mathrm{KL}[p(\bm{\theta}|\mathcal{D}) || q(\bm{\theta})]$, where $p(\bm{\theta}|\mathcal{D})$ was replaced by a Gaussian that had the same mean and covariance as samples drawn from the posterior using the No-U-Turn sampler (NUTS) \cite{hoffman:nuts}, to quantify the calibration of uncertainty estimations.

Results in Fig.~\ref{fig:sep_probit} indicate that EP is the best performing method and that ADF collapses towards a delta function. SEP converges to a solution which appears to be of similar quality to that obtained by EP for the dataset containing Gaussian inputs, but slightly worse when the MoGs was used. Variants of SEP that used larger mini-batches fluctuated less, but typically took longer to converge (although for the small minibatches shown this effect is not clear). The utility of finer grained approximations depended on the homogeneity of the data. For the second dataset containing MoG inputs (shown in Fig.~\ref{fig:daep_probit}), finer-grained approximations were found to be advantageous if the datapoints from each mixture component are assigned to the same approximating factor. Generally it was found that there is no advantage to retaining more approximating factors than there were clusters in the dataset.

To verify whether these conclusions about the granularity of the approximation hold in real datasets, we sampled $N=1,000$ datapoints for each of the digits in MNIST and performed odd-vs-even classification. Each digit class was assigned its own global approximating factor, $K=10$. We compare the log-likelihood of a test set using ADF, SEP ($K=1$), full EP and DSEP ($K=10$) in Figure \ref{fig:mnist}. EP and DSEP significantly outperform ADF. DSEP is slightly worse than full EP initially, however it reduces the memory to 0.001\% of full EP without losing accuracy substantially. SEP's accuracy was still increasing at the end of learning and was slightly better than ADF. Further empirical comparisons are reported in the supplementary, and in summary the three EP methods are indistinguishable when likelihood functions have similar contributions to the posterior.

Finally, we tested SEP's performance on six small binary classification datasets from the UCI machine learning repository.\footnote{\url{https://archive.ics.uci.edu/ml/index.html}} We did not consider the effect of mini-batches or the granularity of the approximation, using $K=M=1$. We ran the tests with damping and stopped learning after convergence (by monitoring the updates of approximating factors). The classification results are summarised in Table \ref{tab:probit_results}. ADF performs reasonably well on the mean classification error metric, presumably because it tends to learn a good approximation to the posterior mode. However, the posterior variance is poorly approximated and therefore ADF returns poor test log-likelihood scores. EP achieves significantly higher test log-likelihood than ADF indicating that a superior approximation to the posterior variance is attained. Crucially, SEP performs very similarly to EP, implying that SEP is an accurate alternative to EP even though it is refining a cheaper global posterior approximation. 

\begin{table} 
\small
\centering  
 \caption{ Average test results all methods on probit regression. All methods appear to capture the posterior's mode, however EP outperforms ADF in terms of test log-likelihood on almost all of the datasets, with SEP performing similarly to EP.}
 \label{tab:probit_results}
\begin{tabular}{l@{\ica}r@{$\pm$}l@{\ica}r@{$\pm$}l@{\ica}r@{$\pm$}l@{\ica}r@{$\pm$}l@{\ica}r@{$\pm$}
	l@{\ica}r@{$\pm$}l@{\ica}r@{$\pm$}}\hline 
{} & \multicolumn{6}{c}{mean error} & \multicolumn{6}{c}{test log-likelihood} \\
\bf{Dataset}&\multicolumn{2}{c}{\bf{ ADF }}&\multicolumn{2}{c}{\bf{ SEP }}&\multicolumn{2}{c}{\bf{ EP }} &\multicolumn{2}{c}{\bf{ ADF }}&\multicolumn{2}{c}{\bf{ SEP }}&\multicolumn{2}{c}{\bf{ EP }} \\ \hline 
Australian&0.328&0.0127&\bf{0.325}&\bf{0.0135}&0.330&0.0133
	&-0.634&0.010&-0.631&0.009&\bf{-0.631}&\bf{0.009}\\
Breast&0.037&0.0045&\bf{0.034}&\bf{0.0034}&0.034&0.0039
	&-0.100&0.015&-0.094&0.011&\bf{-0.093}&\bf{0.011}\\
%
Crabs&0.056&0.0133&\bf{0.033}&\bf{0.0099}&0.036&0.0113
	&-0.242&0.012&-0.125&0.013&\bf{-0.110}&\bf{0.013}\\
%
Ionos&\bf{0.126}&\bf{0.0166}&0.130&0.0147&0.131&0.0149
	&-0.373&0.047&-0.336&0.029&\bf{-0.324}&\bf{0.028}\\
Pima&0.242&0.0093&0.244&0.0098&\bf{0.241}&\bf{0.0093}
	&-0.516&0.013&-0.514&0.012&\bf{-0.513}&\bf{0.012}\\
Sonar&\bf{0.198}&\bf{0.0208}&0.198&0.0217&0.198&0.0243
	&-0.461&0.053&-0.418&0.021&\bf{-0.415}&\bf{0.021}\\
 \hline \end{tabular} 
 \end{table} 
 
 %
\subsection{Mixture of Gaussians for clustering}
The small scale experiments on probit regression indicate that SEP performs well for fully-observed probabilistic models. Although it is not the main focus of the paper, we sought to test the flexibility of the method by applying it to a latent variable model, specifically a mixture of Gaussians. A synthetic MoGs dataset containing $N=200$ datapoints was constructed comprising $J=4$ Gaussians. The means were sampled from a Gaussian distribution, $p(\bm{\mu}_j)= \mathcal{N}(\bm{\mu}; \bm{m}, \mathrm{I})$, the cluster identity variables were sampled from a uniform categorical distribution $p(\bm{h}_n = j) = 1/4$, and each mixture component was isotropic $p(\bm{x}_n | \bm{h}_n) = \mathcal{N}(\bm{x}_n; \bm{\mu}_{\bm{h}_n}, 0.5^2 I)$. EP, ADF and SEP were performed to approximate the joint posterior over the cluster means $\{ \bm{\mu}_j\}$ and cluster identity variables $\{ \bm{h}_n \}$ (the other parameters were assumed known). 

\begin{figure}[ht]
\centering
\def\svgwidth{0.47\linewidth}
\subfigure[\label{fig:gmm_visualised}]{
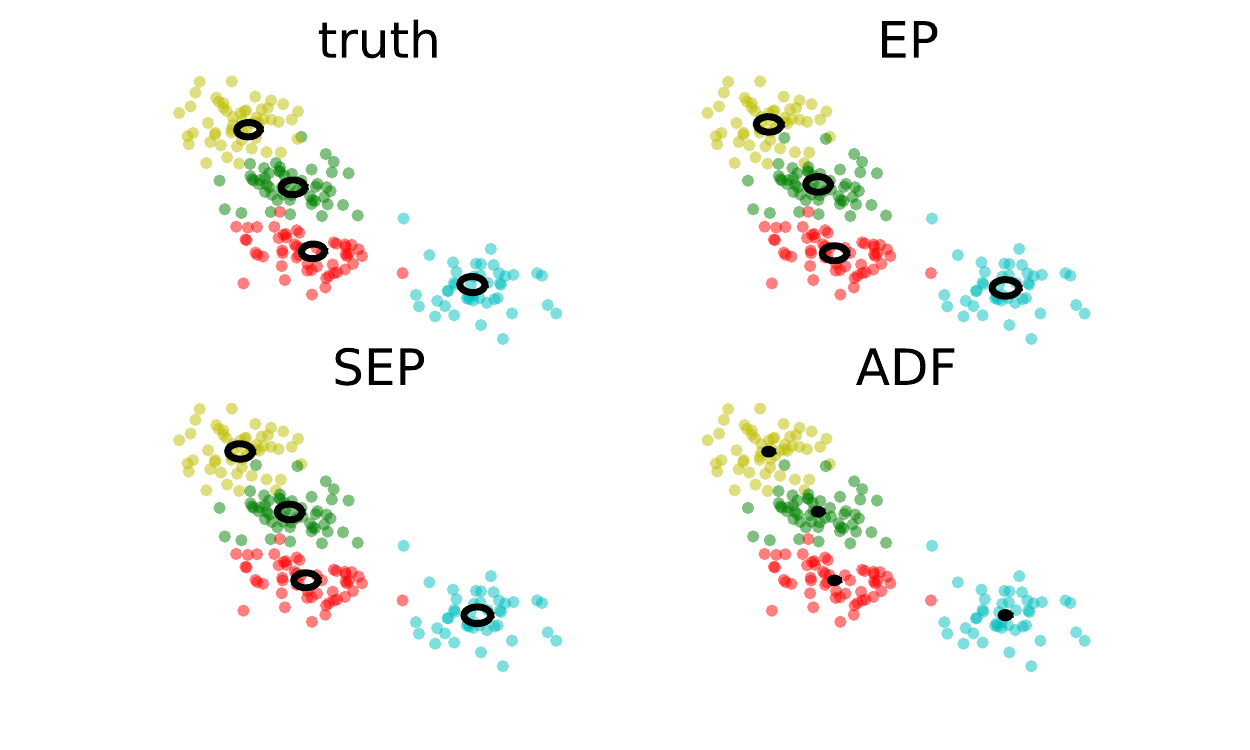}
\hspace{0.1in}
\def\svgwidth{0.40\linewidth}
\subfigure[\label{fig:gmm_error}]{
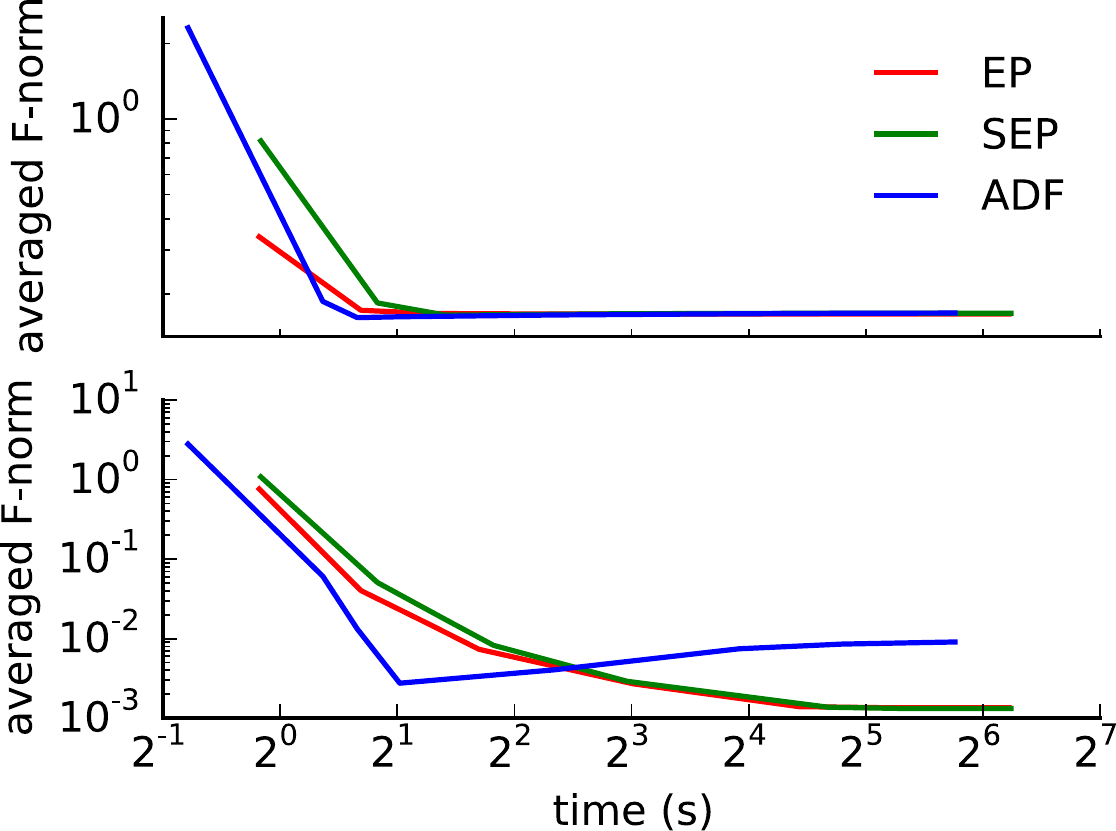}
\caption{Posterior approximation for the mean of the Gaussian components. (a) visualises posterior approximations over the cluster means (98\% confidence level). The coloured dots indicate the true label (top-left) or the inferred cluster assignments (the rest). In (b) we show the error (in F-norm) of the approximate Gaussians' means (top) and covariances (bottom). }
\end{figure}

Figure \ref{fig:gmm_visualised} visualises the approximate posteriors after 200 iterations. All methods return good estimates for the means, but ADF collapses towards a point estimate as expected. SEP, in contrast, captures the uncertainty and returns nearly identical approximations to EP. The accuracy of the methods is quantified in Fig.~\ref{fig:gmm_error} by comparing the approximate posteriors to those obtained from NUTS. In this case the approximate KL-divergence measure is analytically intractable, instead we used the averaged F-norm of the difference of the Gaussian parameters fitted by NUTS and EP methods. These measures confirm that SEP approximates EP well in this case.

\subsection{Probabilistic backpropagation}

The final set of tests consider more complicated models and large datasets. Specifically we evaluate the methods for probabilistic backpropagation (PBP) \cite{miguel:pbp}, a
recent state-of-the-art method for scalable Bayesian learning in neural
network models. Previous implementations of PBP perform several iterations of ADF over the training
data.  The moment matching operations required by ADF are themselves intractable and they are approximated by
first propagating the uncertainty on the synaptic weights forward through the network in a sequential way, 
and then computing the gradient of the marginal likelihood by backpropagation.
ADF is used to reduce the large memory cost that would be required by EP
when the amount of available data is very large.

We performed several experiments to assess the accuracy of different implementations of PBP based on ADF, SEP and EP on regression datasets following the same experimental protocol as in \cite{miguel:pbp} (see supplementary material). 
We considered neural networks with 50 hidden units (except for \textit{Year} and \textit{Protein} which we used 100).
Table \ref{tab:pbp_results} shows the average test RMSE and test log-likelihood for each method. Interestingly, SEP can outperform EP in this setting (possibly because the stochasticity enabled it to find better solutions), and typically it performed similarly. Memory reductions using SEP instead of EP were large e.g.~694Mb for the Protein dataset and 65,107Mb for the Year dataset (see supplementary). Surprisingly ADF often outperformed EP, although the results presented for ADF use a near-optimal number of sweeps and further iterations generally degraded performance. ADF's good performance is most likely due to an interaction with additional moment approximation required in PBP that is more accurate as the number of factors increases.

\begin{table} 
\small
\centering 
\caption{Average test results for all methods. Datasets are also from the UCI machine learning repository.}
\label{tab:pbp_results}
\begin{tabular}{l@{\ica}r@{$\pm$}l@{\ica}r@{$\pm$}l@{\ica}r@{$\pm$}l@{\ica}r@{$\pm$}l@{\ica}r@{$\pm$}l@{\ica}r@{$\pm$}l@{\ica}r@{$\pm$}}\hline 
{} & \multicolumn{6}{c}{RMSE} & \multicolumn{6}{c}{test log-likelihood} \\
\bf{Dataset}&\multicolumn{2}{c}{\bf{ ADF }}&\multicolumn{2}{c}{\bf{ SEP }}&\multicolumn{2}{c}{\bf{ EP }} &\multicolumn{2}{c}{\bf{ ADF }}&\multicolumn{2}{c}{\bf{ SEP }}&\multicolumn{2}{c}{\bf{ EP }} \\ \hline 
Kin8nm&0.098&0.0007&\bf{0.088}&\bf{0.0009}&0.089&0.0006
	&0.896&0.006&\bf{1.013}&\bf{0.011}&1.005&0.007\\ 
Naval&0.006&0.0000&\bf{0.002}&\bf{0.0000}&0.004&0.0000
	&3.731&0.006&\bf{4.590}&\bf{0.014}&4.207&0.011\\  
Power&\bf{4.124}&\bf{0.0345}&4.165&0.0336&4.191&0.0349
	&\bf{-2.837}&\bf{0.009}&-2.846&0.008&-2.852&0.008\\
Protein&4.727&0.0112&\bf{4.670}&\bf{0.0109}&4.748&0.0137
	&-2.973&0.003&\bf{-2.961}&\bf{0.003}&-2.979&0.003\\ 
Wine&\bf{0.635}&\bf{0.0079}&0.650&0.0082&0.637&0.0076
	&-0.968&0.014&-0.976&0.013&\bf{-0.958}&\bf{0.011}\\  
Year&\bf{8.879}&\bf{   NA}&8.922&   NA&8.914&   NA
&\bf{-3.603}&\bf{  NA}&-3.924&  NA&-3.929&  NA\\
 \hline \end{tabular} 
 \end{table}

\section{Conclusions and future work}
This paper has presented the stochastic expectation propagation method for reducing EP's large memory consumption which is prohibitive for large datasets. We have connected the new algorithm to a number of existing methods including assumed density filtering, variational message passing, variational inference, stochastic variational inference and averaged EP.
Experiments on Bayesian logistic regression (both synthetic and real world) and mixture of Gaussians clustering indicated that the new method had an accuracy that was competitive with EP.  Experiments on the probabilistic back-propagation on large real world regression datasets again showed that SEP comparably to EP with a vastly reduced memory footprint. 
Future experimental work will focus on developing data-partitioning methods to leverage finer-grained approximations (DESP) that showed promising experimental performance and also mini-batch updates. There is also a need for further theoretical understanding of these algorithms, and indeed EP itself. Theoretical work will study the convergence properties of the new algorithms for which we only have limited results at present. Systematic comparisons of EP-like algorithms and variational methods will guide practitioners to choosing the appropriate scheme for their application.

\subsection*{Acknowledgements}
We thank the reviewers for valuable comments. YL thanks the Schlumberger Foundation Faculty for the Future fellowship on supporting her PhD study. JMHL acknowledges support from the Rafael del Pino Foundation. RET thanks EPSRC grant \# EP/G050821/1 and EP/L000776/1.

\subsubsection*{References}
\renewcommand{\section}[2]{}
\small
\bibliographystyle{unsrt}
\bibliography{nips_sep}



\section*{Supplementary material}

# Stochastic Expectation Propagation: Supplementary Material

**Yingzhen Li**
University of Cambridge
Cambridge, CB2 1PZ, UK
`yl494@cam.ac.uk`

**José Miguel Hernández-Lobato**
Harvard University
Cambridge, MA 02138 USA
`jmh@seas.harvard.edu`

**Richard E. Turner**
University of Cambridge
Cambridge, CB2 1PZ, UK
`ret26@cam.ac.uk`

The supplementary material is divided into these sections. Section A details the design of stochastic power EP methods and presents relationships between SEP and SVI. Section B extends the discussion of distributed algorithms and SEP's applicability to latent variable models. Section C provides experimental details of the Bayesian neural network experiments and presents further emprical evaluations of the method.

## A  Further theoretical connections

We described the extensions of stochastic expectation propagation (SEP) in the main text, and we provide more details in this section.

### A.1  Power EP and alpha-EP

The relationship between EP and variational inference (VI) can be explained by introducing power EP (PEP) [1]. As a preparation let us consider the alpha-divergence[1] introduced in [2]

$$D_\alpha[p(\boldsymbol{\theta})||q(\boldsymbol{\theta})] = \frac{4}{1-\alpha^2} \left(1 - \int_{\boldsymbol{\theta}} p(\boldsymbol{\theta})^{\frac{1+\alpha}{2}} q(\boldsymbol{\theta})^{\frac{1-\alpha}{2}} d\boldsymbol{\theta} \right). \tag{1}$$

Two cases of KL-divergence also belongs to the family of alpha-divergence by definition:

$$D_1[p(\boldsymbol{\theta})||q(\boldsymbol{\theta})] \stackrel{\triangle}{=} \lim_{\alpha \to 1} D_\alpha[p(\boldsymbol{\theta})||q(\boldsymbol{\theta})] = \mathrm{KL}[p(\boldsymbol{\theta})||q(\boldsymbol{\theta})], \tag{2}$$

$$D_{-1}[p(\boldsymbol{\theta})||q(\boldsymbol{\theta})] \stackrel{\triangle}{=} \lim_{\alpha \to -1} D_\alpha[p(\boldsymbol{\theta})||q(\boldsymbol{\theta})] = \mathrm{KL}[q(\boldsymbol{\theta})||p(\boldsymbol{\theta})]. \tag{3}$$

Minka [1] also introduced alpha-EP as a generalisation of EP to alpha-divergences, which changes the moment matching step to alpha-projection [3] that returns the minimiser of the alpha divergence $D_\alpha[\tilde{p}_n(\boldsymbol{\theta})||q(\boldsymbol{\theta})]$ wrt. $q(\boldsymbol{\theta})$. Examples include moment projection `proj`[·] which takes $\alpha = 1$, and information projection which chooses $\alpha = -1$. However alpha-projections are difficult to compute in general, motivating power EP (Algorithm 1) – so called because it raises potentials to a power before referencing standard EP updates – as a practical alternative. Minka [1] showed that power EP with power $1/\beta, \beta < \infty$ returns a local optimum of the alpha divergence with $\alpha = -1 + 2/\beta$ when converged. However this still leaves the pathological case $\alpha = -1$ or $\beta = \infty$ since the above equivalence does not apply. Thus variational message passing (VMP), which takes $\alpha \to -1$, cannot be interpreted as a special case of power EP. This observation extends to stochastic PEP as well (Algorithm 2). Instead we derive stochastic VMP in the spirit which SEP extends EP, which keeps the computational steps using current global estimate but ties all the local factors. We discuss this extension in detail in the next section and provide its connection to stochastic variational inference.

---

[1] A little math can show the updates of alpha-EP using different existing alpha-divergence definitions are equivalent, although the corresponding alpha will change.

| **Algorithm 1** PEP | **Algorithm 2** Stochastic PEP |
|---|---|
| 1: choose a factor $f_n$ to refine: | 1: choose a datapoint $\boldsymbol{x}_n \sim \mathcal{D}$: |
| 2: compute cavity distribution | 2: compute cavity distribution |
| $\qquad q_{-n}(\boldsymbol{\theta}) \propto q(\boldsymbol{\theta})/f_n(\boldsymbol{\theta})^{1/\beta}$ | $\qquad q_{-1}(\boldsymbol{\theta}) \propto q(\boldsymbol{\theta})/f(\boldsymbol{\theta})^{1/\beta}$ |
| 3: compute tilted distribution | 3: compute tilted distribution |
| $\qquad \tilde{p}_n(\boldsymbol{\theta}) \propto p(\boldsymbol{x}_n|\boldsymbol{\theta})^{1/\beta} q_{-n}(\boldsymbol{\theta})$ | $\qquad \tilde{p}_n(\boldsymbol{\theta}) \propto p(\boldsymbol{x}_n|\boldsymbol{\theta})^{1/\beta} q_{-1}(\boldsymbol{\theta})$ |
| 4: moment matching: | 4: moment matching: |
| $\qquad f_n(\boldsymbol{\theta}) \leftarrow [\texttt{proj}[\tilde{p}_n(\boldsymbol{\theta})]/q_{-n}(\boldsymbol{\theta})]^{\beta}$ | $\qquad f_n(\boldsymbol{\theta}) \leftarrow [\texttt{proj}[\tilde{p}_n(\boldsymbol{\theta})]/q_{-1}(\boldsymbol{\theta})]^{\beta}$ |
| 5: inclusion: | 5: inclusion: |
| $\qquad q(\boldsymbol{\theta}) \leftarrow q(\boldsymbol{\theta})f_n(\boldsymbol{\theta})/f_n^{old}(\boldsymbol{\theta})$ | $\qquad q(\boldsymbol{\theta}) \leftarrow q(\boldsymbol{\theta})f_n(\boldsymbol{\theta})/f(\boldsymbol{\theta})$ |
| | 6: *implicit update*: |
| | $\qquad f(\boldsymbol{\theta}) \leftarrow f(\boldsymbol{\theta})^{1-\frac{1}{N}} f_n(\boldsymbol{\theta})^{\frac{1}{N}}$ |

## A.2 Connecting SVMP to SVI

We first briefly sketch the VMP algorithm using the EP framework, but replacing the moment matching step with natural parameter matching. We assume the approximate posterior $q(\boldsymbol{\theta})$ is in some exponential family:

$$q(\boldsymbol{\theta}) \propto \exp\left[\langle \boldsymbol{\lambda}_q, \boldsymbol{\phi}(\boldsymbol{\theta})\rangle\right]. \tag{4}$$

At time $t$ we have the current estimate of the natural parameter $\boldsymbol{\lambda}_q^t$, which is defined as the sum of local variational parameters[2]: $\boldsymbol{\lambda}_q^t \triangleq \boldsymbol{\lambda}_0 + \sum_{n=1}^{N} \boldsymbol{\lambda}_n^t$. Here $\boldsymbol{\lambda}_0$ represents the natural parameter of the prior distribution $p_0(\boldsymbol{\theta})$. VMP iteratively computes the update of each local estimate $\boldsymbol{\lambda}_n^{t+1}$ in the following procedure. First VMP computes the expected sufficient statistics $\hat{\boldsymbol{s}}_n$ about datapoint $\boldsymbol{x}_n$ using $\boldsymbol{\lambda}_q^t$, e.g. $\hat{\boldsymbol{s}}_n = E_q[t(z_n, x_n)]$ in the original SVI paper [4]. Then VMP forms the gradient as though optimising the maximised evidence lower bound (ELBO) but with $q_{-1}(\boldsymbol{\theta})$ as the prior:

$$\nabla_{\boldsymbol{\lambda}_q^t}\mathcal{L} = \boldsymbol{\lambda}_{-1}^t + \hat{\boldsymbol{s}}_n - \boldsymbol{\lambda}_q^t, \tag{5}$$

$$\boldsymbol{\lambda}_{-1}^t = \boldsymbol{\lambda}_q^t - \boldsymbol{\lambda}_n^t. \tag{6}$$

Next VMP zeros the gradient and recovers the current update $\boldsymbol{\lambda}_n^{t+1} = \hat{\boldsymbol{s}}_n$. The stochastic version of VMP, if extended in a way as SEP developed from EP, defines the global variational parameters as $\boldsymbol{\lambda}_q^t \triangleq \boldsymbol{\lambda}_0 + N\boldsymbol{\lambda}^t$. It computes the expected sufficient statistics $\hat{\boldsymbol{s}}_n$ in the same way as VMP but changes the cavity to $\boldsymbol{\lambda}_{-1}^t = \boldsymbol{\lambda}_q^t - \boldsymbol{\lambda}^t$ in the ELBO maximisation steps. Readers can verify that this returns the current update $\boldsymbol{\lambda}^{t+1} = \hat{\boldsymbol{s}}_n$ using the important fact that the local update ONLY depends on the global parameter $\boldsymbol{\lambda}_q^t$. Now since we tie all the local updates, the global parameter update $\boldsymbol{\lambda}_q^{t+1} = \boldsymbol{\lambda}_0 + N\boldsymbol{\lambda}^{t+1} = \boldsymbol{\lambda}_0 + N\hat{\boldsymbol{s}}_n$. In practice we perform a damped update, where a typical choice of step size is $\epsilon = 1/N$ like in SEP:

$$\boldsymbol{\lambda}_q^{t+1} \leftarrow (1 - \frac{1}{N})\boldsymbol{\lambda}_q^t + \frac{1}{N}(\boldsymbol{\lambda}_0 + N\boldsymbol{\lambda}^{t+1}) = \boldsymbol{\lambda}_0 + (N-1)\boldsymbol{\lambda}^t + \hat{\boldsymbol{s}}_n. \tag{7}$$

On the other hand, [5] summarises stochastic variational inference (SVI) as to compute the current update by zeroing the gradient

$$\nabla_{\boldsymbol{\lambda}_q}\mathcal{L} = \boldsymbol{\lambda}_0 + N\hat{\boldsymbol{s}}_n - \boldsymbol{\lambda}_q, \tag{8}$$

which returns $\boldsymbol{\lambda}_q^{t+1} = \boldsymbol{\lambda}_0 + N\hat{\boldsymbol{s}}_n$ as well. This implies that SVI, when using learning rate $\epsilon = 1/N$, is equivalent to SVMP.

## A.3 SEP from a global approximation perspective

In this section we provide some intuition about SEP via an interpretation as approximating minimisation of a global divergence like VI (although it is computed in a truly local way). This framework utilises alpha-divergence, but on the global posterior, and we motivate it by describing VI and SVI as

---

[2]This notation implicitly assumes that the prior and the approximate posterior belong to the approximate distribution family. In general we can propose another factor to approximate $p_0(\boldsymbol{\theta})$, and our result still applies.

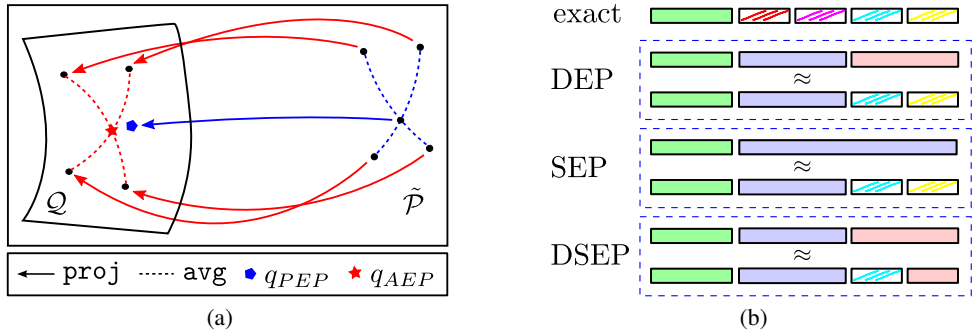

(a)                                      (b)

Figure 1: (a) A geometric view of AEP/PEP comparison. (b) A cartoon illustration of DEP, SEP and DSEP. For each algorithm we show the approximate posterior on the top and the tilted distribution at the bottom.

divergence minimisation. VI performs global optimisation on $\text{KL}[q(\boldsymbol{\theta})||p(\boldsymbol{\theta}|\mathcal{D})]$, and its stochastic version, SVI, can be interpreted as at each step minimising $\text{KL}[q(\boldsymbol{\theta})||p(\boldsymbol{\theta}|\{\boldsymbol{x}_n\}^N)]$ with the $N$ replicas $\{\boldsymbol{x}_n\}^N = \{\boldsymbol{x}_n, \boldsymbol{x}_n, ..., \boldsymbol{x}_n\}$. Similarly, we state SEP as a stochastic global optimisation procedure, which computes an iterative procedure to minimise alpha-divergence $D_\alpha[p(\boldsymbol{\theta}|\{\boldsymbol{x}_n\}^N)||q(\boldsymbol{\theta})]$ with $\alpha = -1 + 2/N$. Indeed we can understand the inner-loop of AEP as PEP with power $1/N$ if considering $f(\boldsymbol{\theta})^N$ as a large composite factor to approximate the likelihood term of $\boldsymbol{x}_n$ raised to power $N$.

However minimising the alpha-divergence between the true posterior $p(\boldsymbol{\theta}|\mathcal{D})$ and the global approximation $q(\boldsymbol{\theta})$ recovers PEP on the whole dataset instead, and the factor to include in the tilted distribution changes to the intractable geometric average $\text{avg}[\{p(\boldsymbol{x}_n|\boldsymbol{\theta})\}] \overset{\triangle}{=} [\prod_n p(\boldsymbol{x}_n|\boldsymbol{\theta})]^{1/N}$. Readers might have noticed that the update of PEP on the full dataset is given by $q(\boldsymbol{\theta}) \leftarrow \texttt{proj}[\text{avg}[\{\tilde{p}_n(\boldsymbol{\theta})\}]]$. In other words, we can interpret AEP as an approximation to the impractical batch PEP by interchanging projections and averaging operations, and we illustrate a geometric view for this in Fig. 1(a).

It is important to note that SEP/AEP at convergence does not minimise the alpha divergence globally. Like PEP, the inner-loop computes $\texttt{proj}[\tilde{p}_n(\boldsymbol{\theta})]$, where one can show that it moves towards minimising $D_\alpha(p(\boldsymbol{\theta}|\{\boldsymbol{x}_n\}^N)||q_n(\boldsymbol{\theta}))$ using the same techniques as before. However the outer-loop averages the natural parameters of the intermediate answers, which does not follow the gradient direction of alpha-divergence minimisation. Furthermore, local/global optimisation of alpha divergence are inconsistent in terms of fixed points except at $\alpha = -1$, the divergence utilised in VI and VMP. Indeed we provide the fixed point conditions of AEP which reveals its local nature.

**Proposition 1.** *The fixed points of averaged EP, if they exist, can be written as $q(\boldsymbol{\theta}) = \texttt{avg}[\{q_n(\boldsymbol{\theta})\}]$, where*

$$q_n(\boldsymbol{\theta}) = \texttt{proj}[\tilde{p}_n(\boldsymbol{\theta})], \tag{9}$$

$$\tilde{p}_n(\boldsymbol{\theta}) \propto q(\boldsymbol{\theta})\frac{p(\boldsymbol{x}_n|\boldsymbol{\theta})}{f(\boldsymbol{\theta})}. \tag{10}$$

*These fixed points are also the fixed points of stochastic EP when the learning rates satisfy the Robbins-Monro condition [6].*

This fixed point condition applies to stochastic PEP as well when $\alpha \neq -1$, and importantly it also implies the pathology of constructing SVMP by using SPEP and limiting $\alpha$ to $-1$.

## B  Algorithmic design details

### B.1  Distributed computing methods

We have shown in the main paper that a proper design of data partitioning improves SEP's approximation accuracy. This distributed algorithm is inspired by the Distributed EP (DEP) algorithm [7, 8]

| **Algorithm 3** DEP | **Algorithm 4** DSEP | **Algorithm 5** DAEP |
|---|---|---|
| 1: compute cavity distribution $q_{-k}(\boldsymbol{\theta}) \propto q(\boldsymbol{\theta})/f_k(\boldsymbol{\theta})$ | 1: compute cavity distribution $q_{-1}(\boldsymbol{\theta}) = q(\boldsymbol{\theta})/f_k(\boldsymbol{\theta})$ | 1: compute cavity distribution $q_{-1}(\boldsymbol{\theta}) \propto q(\boldsymbol{\theta})/f_k(\boldsymbol{\theta})$ |
| 2: compute tilted distribution $\tilde{p}_k(\boldsymbol{\theta}) \propto p(\mathcal{D}_k|\boldsymbol{\theta})q_{-k}(\boldsymbol{\theta})$ | 2: choose a datapoint $\boldsymbol{x}_n \sim \mathcal{D}_k$ | 2: for each $\boldsymbol{x}_n \in \mathcal{D}_k$: |
| 3: moment matching: $f_k(\boldsymbol{\theta}) \leftarrow \texttt{proj}[\tilde{p}_k(\boldsymbol{\theta})]/q_{-k}(\boldsymbol{\theta})$ | 3: compute tilted distribution $\tilde{p}_k^n(\boldsymbol{\theta}) \propto p(\boldsymbol{x}_n|\boldsymbol{\theta})q_{-1}(\boldsymbol{\theta})$ | 3:    compute tilted distribution $\tilde{p}_k^n(\boldsymbol{\theta}) \propto p(\boldsymbol{x}_n|\boldsymbol{\theta})q_{-1}(\boldsymbol{\theta})$ |
| | 4: moment matching: $f_k^n(\boldsymbol{\theta}) \leftarrow \texttt{proj}[\tilde{p}_k^n(\boldsymbol{\theta})]/q_{-1}(\boldsymbol{\theta})$ | 4:    moment matching: $f_k^n(\boldsymbol{\theta}) \leftarrow \texttt{proj}[\tilde{p}_k^n(\boldsymbol{\theta})]/q_{-1}(\boldsymbol{\theta})$ |
| | 5: inclusion: $f_k(\boldsymbol{\theta}) \leftarrow f_k(\boldsymbol{\theta})^{1-1/N_k} f_k^n(\boldsymbol{\theta})^{1/N_k}$ | 5: inclusion: $f_k(\boldsymbol{\theta})^{N_k} \leftarrow \prod_n f_k^n(\boldsymbol{\theta})$ |

Figure 2: Comparing the variants of distributed design for Expectation Propagation (EP) on the current data piece $\mathcal{D}_k$. One should notice that the definitions of $f_k(\boldsymbol{\theta})$ are different for DEP and DSEP/DAEP. Distributed EP (DEP) uses sampling methods to compute the projection step, while Distributed SEP/AEP (DSEP/DAEP) still keeps deterministic computations.

presented in Algorithm 3. DEP first partitions the dataset into $K$ disjoint pieces $\{D_k = \{\boldsymbol{x}_i\}_{i=1}^{N_k}\}$ with $N = \sum_{i=1}^{K} N_k$, which is well-justified since the true posterior can also be derived as

$$p(\boldsymbol{\theta}|\mathcal{D}) \propto p_0(\boldsymbol{\theta}) \prod_{k=1}^{K} p(\mathcal{D}_k|\boldsymbol{\theta}), \tag{11}$$

$$p(\mathcal{D}_k|\boldsymbol{\theta}) = \prod_{\boldsymbol{x}_n \in \mathcal{D}_k} p(\boldsymbol{x}_n|\boldsymbol{\theta}). \tag{12}$$

Next DEP assigns factors to each sub-dataset likelihood, i.e. $q(\boldsymbol{\theta}) \propto p_0(\boldsymbol{\theta}) \prod_k f_k(\boldsymbol{\theta})$ with each $f_k(\boldsymbol{\theta})$ approximating $p(\mathcal{D}_k|\boldsymbol{\theta})$. The projection step is no longer analytically tractable in general since the tilted distribution with multiple datapoints often lacks a simple form. Instead DEP handles moment matching with sampling, making it stochastic in the sense of having an stochastic approximation of the moments.

To construct a deterministic counterpart of DEP, we consider running SEP/AEP **inside** each partition. We name this approach as Distributed SEP/AEP (DSEP/DAEP) and provide a comparison in Fig. 1(b) with DEP and SEP **on** the sub-dataset likelihood factors using sampling protocol. Different from DEP, the approximate posterior for DSEP is defined as $q(\boldsymbol{\theta}) \propto p_0(\boldsymbol{\theta}) \prod_k f_k(\boldsymbol{\theta})^{N_k}$, with $f_k(\boldsymbol{\theta})^{N_k}$ approximating $p(\mathcal{D}_k|\boldsymbol{\theta})$. The computations are almost the same as SEP/AEP except that the updates only modify the copies of the corresponded subset. These two algorithms are also detailed in Algorithm 4 and 5, respectively. In section C.3 we provide an empirical study on comparing SEP, EP and DSEP approximations.

## B.2 SEP with latent variables

In this section we show the applicability of SEP to latent variables without scaling the memory consumption with $N$. We consider a model containing latent variables $\boldsymbol{h}_n$ associated with each observation $\boldsymbol{x}_n$, which are drawn i.i.d. from a prior $p_0(\boldsymbol{h}_n)$. SEP proposes approximations to the true posterior over parameters and hidden variables

$$p(\boldsymbol{\theta}, \{\boldsymbol{h}_n\}|\mathcal{D}) \propto p_0(\boldsymbol{\theta}) \prod_n p_0(\boldsymbol{h}_n) p(\boldsymbol{x}_n|\boldsymbol{h}_n, \boldsymbol{\theta}) \tag{13}$$

by tying the factors for the global parameter $\boldsymbol{\theta}$ but retaining the local factors for the hidden variables:

$$q(\boldsymbol{\theta}, \{\boldsymbol{h}_n\}) \overset{\triangle}{\propto} p_0(\boldsymbol{\theta}) f(\boldsymbol{\theta})^N \prod_{n=1}^{N} g_n(\boldsymbol{h}_n). \tag{14}$$

In other words, SEP uses $f(\boldsymbol{\theta})g_n(\boldsymbol{h}_n)$ to approximate $p(\boldsymbol{x}_n|\boldsymbol{h}_n, \boldsymbol{\theta})p_0(\boldsymbol{h}_n)$.

Next we show a critical advantage of SEP when approximating the latent variable posterior distributions: the local factors $g_n(\boldsymbol{h}_n)$ do not need to be maintained in memory (though see caveats mentioned below). More formally, the cavity distribution is $q_{-n}(\boldsymbol{\theta}, \{\boldsymbol{h}_n\}) \propto q(\boldsymbol{\theta}, \{\boldsymbol{h}_n\})/(f(\boldsymbol{\theta})g_n(\boldsymbol{h}_n))$

Table 1: Datasets Used in the Experiments with Neural Networks. The memory numbers reported include dataset storage and temporal maintainance of computation graphs in Theano ($\sim 100 MB$).

| Dataset | $N$ | $D$ | MB (EP) | MB (SEP) | MB reduced |
|---------|-----|-----|---------|----------|------------|
| Kin8nm | 8192 | 8 | 168.23 | 109.76 | 58.47 |
| Naval Propulsion | 11,934 | 16 | 261.75 | 113.92 | 147.83 |
| Combined Cycle Power Plant | 9568 | 4 | 148.70 | 110.99 | 37.71 |
| Protein Structure | 45,730 | 9 | 815.55 | 121.52 | 694.02 |
| Wine Quality Red | 1599 | 11 | 122.21 | 107.90 | 14.30 |
| **Year Prediction MSD** | **515,345** | **90** | **67837.90** | **2730.55** | **65107.34** |

and the tilted distribution is $\tilde{p}_n(\boldsymbol{\theta}, \{\boldsymbol{h}_n\}) \propto q_{-n}(\boldsymbol{\theta}, \{\boldsymbol{h}_n\})p(\boldsymbol{x}_n|\boldsymbol{h}_n, \boldsymbol{\theta})p_0(\boldsymbol{h}_n)$. This leads to the a moment-update that minimises

$$\text{KL}\left[p_0(\boldsymbol{\theta})f(\boldsymbol{\theta})^{N-1}p(\boldsymbol{x}_n|\boldsymbol{h}_n, \boldsymbol{\theta})p_0(\boldsymbol{h}_n)\prod_{m \neq n} g_m(\boldsymbol{h}_m)||p_0(\boldsymbol{\theta})f(\boldsymbol{\theta})^{N-1}f'(\boldsymbol{\theta})g_n(\boldsymbol{h}_n)\prod_{m \neq n} g_m(\boldsymbol{h}_m)\right].$$

with respect to $f'(\boldsymbol{\theta})g_n(\boldsymbol{h}_n)$. Importantly, the terms involving $\prod_{m \neq n} g_m(\boldsymbol{h}_m)$ cancel, meaning that these factors do not contribute to the local approximation step. For simple models the moments of $\boldsymbol{h}_n$ can be computed analytically given $q_{-1}(\boldsymbol{\theta})$, thus $g_n(\boldsymbol{h}_n)$ is never stored in memory resulting in a reduced memory footprint by a factor of $N$ again. However in practice people may prefer maintaining the $g$ factors in memory, if the moment computation requires another optimisation inner-loop (which might be more expensive than the moment matching step itself). Examples include latent Dirichlet allocation [9] that has a hierachy of latent variables, where VI methods also store variational $q$ distributions for some of the hidden variables. One potential recipe in this scenario is to learn the moments/messages passed in each SEP step in the spirit of [10, 11].

It is also possible to have latent variables globally shared or shared in a data subset $\mathcal{D}_k$. But we can also extend SEP to these latent variables accordingly, which still provides computation gains in space complexity. In mathematical forms, assume $\boldsymbol{h}_k$ a latent variable shared in $\mathcal{D}_k$. Then we construct $q(\boldsymbol{h}_k) \propto p_0(\boldsymbol{h}_k)g_k(\boldsymbol{h}_k)^{N_k}$ to approximate its posterior. This procedure still reduces memory by a factor of $N/K$.

# C    Further experimental results

## C.1    Details of Bayesian neural network tests

We perform neural network regression experiments with publicly available data sets and neural networks with one hidden layer. Table 1 lists the analyzed data sets and shows summary statistics. We use neural networks with 50 hidden units in all cases except in the two largest ones, i.e., *Year Prediction MSD* and *Protein Structure*, where we use 100 hidden units. The different methods, SEP, EP and ADF were run by performing 40 passes over the available training data, updating the parameters of the posterior approximation after seeing each data point. The data sets are split into random training and test sets with 90% and 10% of the data, respectively. This splitting process is repeated 20 times and the average test performance of each method is reported. In the two largest data sets, *Year Prediction MSD* and *Protein Structure*, we do the train-test splitting only once and five times respectively. The data sets are normalized so that the input features and the targets have zero mean and unit variance in the training set. The normalization on the targets is removed for prediction.

We also provide the memory consumption details for experiments using probabilistic back-propagation (PBP) in Table 1. We observe substantial memory reductions by running SEP instead of EP, while still attaining similar accuracies. Especially for Year Prediction MSD dataset, which is a typical large-scale dataset both in the number of observations $N$ and the dimensionality $D$, SEP achieves saving tens of gigabytes. We performed the test for EP using a machine with more than 100GB RAM, while SEP only required 2.7GB memory, including the space of storing the dataset (1.9GB). These numbers reveal the huge memory requirement of full EP and further support SEP as a practical alternative in big data, big model settings.

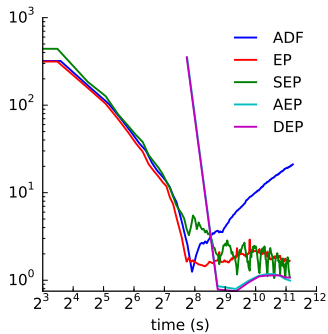

Figure 3: Performance of EP methods on Bayesian logistic regression with sampling moment computations, measured in approximate KL divergence described in the main text.

### C.2 Stochastic EP with sampling protocal

Although not a main purpose, we further test the performance of SEP when using sampling methods to compute moments [3]. We re-use the settings of probit regression but change the probit unit to sigmoid function, making the moment projection analytically intractable. We randomly partition the dataset into $K = 20$ subsets $\{\mathcal{D}_k\}$, construct the approximate posterior with local factors over the subsets, and tie them in SEP/AEP as before. Note that we perform sequential computations for DEP and AEP although they are ideally suited for parallel computing. Again as presented in Figure 3, SEP performs almost as well as EP, which further justifies SEP even with sampling methods. Also AEP is indistinguishable from DEP, but it reduces memory by a factor of $K$.

### C.3 Further Comparisons for SEP, DSEP and full EP

The assumption we made in the main text to achieve SEP ≈ full EP is that the contributions of each likelihood term to the posterior are similar. We show further results here on the approximation produced by different EP methods when there is significant heterogeneity in the data. We generated synthetic XOR classification data by sampling from 4 unit Gaussians with means $(3, 3)$, $(-3, -3)$, $(3, -3)$ and $(-3, 3)$, and labelling the clusters centered at the former two as negative examples (and positive for the others). The model $p(y_n|\boldsymbol{x}_n, \boldsymbol{\theta})$ is kernel probit regression using RBF kernel with width $l = 1.0$, which is the same as the model presented in Section 5.1 in the main text except that the features are changed to kernel representations. This makes the feature vectors high dimensional, and the local nature of kernels also makes the kernel-expanded inputs very different if the datapoints belong to different clusters. We generated $50 * 4$ test data and $\{10 * 4, 20 * 4, 50 * 4\}$ training data and ran SEP/DSEP/full EP to approximate the posterior distribution of $\boldsymbol{\theta}$. For DSEP we partitioned the dataset into 4 subsets according to the associated centroid. Each experiment was repeated 10 times to collect average test data log-likilihood and classification errors.

Table 2 shows the quatitative numbers of performances and Figure 4 visualises the contours of probability $p(y = 1|\boldsymbol{x}, \mathcal{D})$ with true posterior approximated by $q(\boldsymbol{\theta})$. Interestingly SEP is slightly better then the others on the classification error metric. But importantly EP achieves the best test log-likelihood numbers and in general DSEP produces very similar results (shown by both the table and the figure), meaning that even for small datasets running full EP might be unnecessary. Also the three methods become indistinguishable when the size of the dataset $N$ increases. We argue the main reason is that the posterior contributions are getting similar since more datapoints are observed in the circle of kernel width.

We further tested the robustness of all three methods to outliers. We reused the settings above and randomly flipped $10\%$ labels of training data. Qualitative results in Figure 5 show that SEP is almost as robust as DSEP/EP in this example. We had tried different types of outliers and failed to find the cases where EP/DSEP significantly outperforms SEP. Future work should further characterises that when SEP gives bad approximations and separately whether it fails in the same way as EP fails, e.g. EP can fail to converge.

---

[3]code adjusted from `ep-stan`: https://github.com/gelman/ep-stan

Table 2: Average test results of all methods on kernel Probit regression.

| $N$ | mean error | | | test log-likelihood | | |
| | **SEP** | **DSEP** | **EP** | **SEP** | **DSEP** | **EP** |
|---|---|---|---|---|---|---|
| $10*4$ | **0.032±0.0058** | 0.055±0.0127 | 0.032±0.0097 | -0.405±0.011 | -0.380±0.010 | **-0.378±0.009** |
| $20*4$ | **0.007±0.0014** | 0.008±0.0024 | 0.012±0.0031 | -0.326±0.007 | -0.320±0.006 | **-0.317±0.003** |
| $50*4$ | **0.003±0.0010** | 0.003±0.0014 | 0.006±0.0009 | -0.243±0.004 | **-0.233±0.007** | -0.238±0.003 |

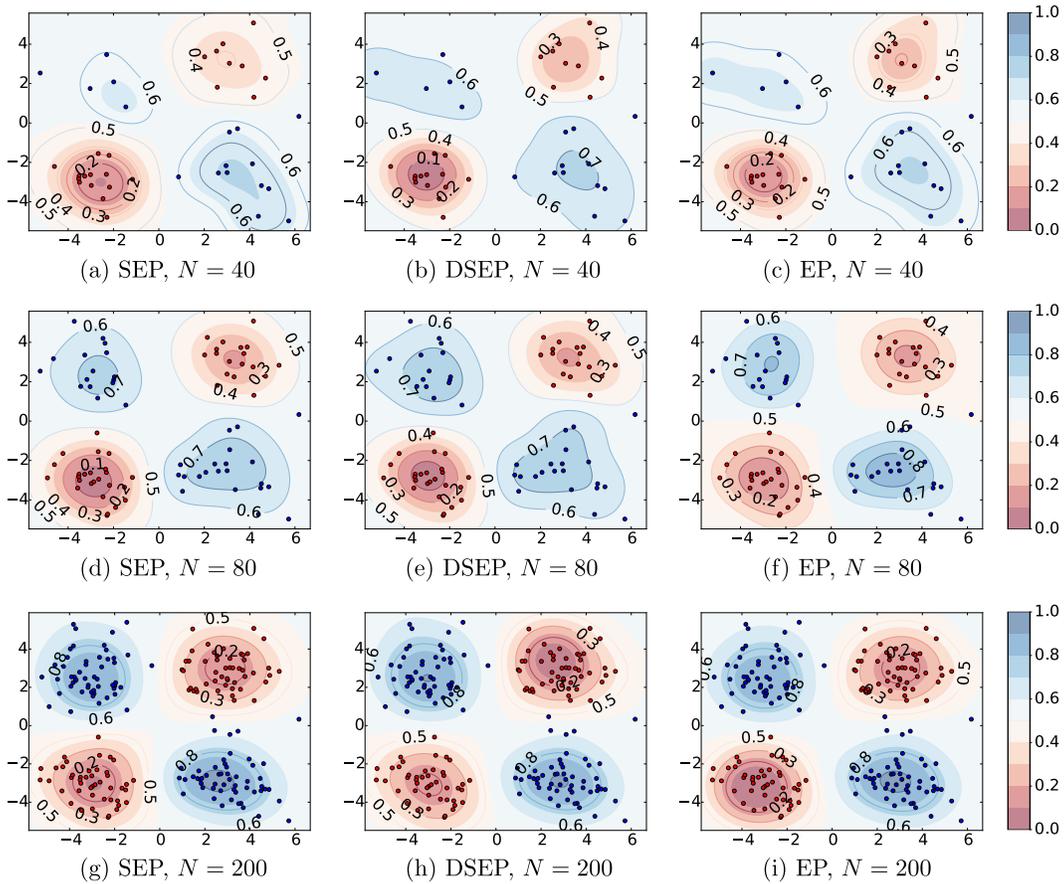

Figure 4: Comparing predictions of kernel probit regression trained by SEP/DSEP/EP, with increasing training data size $N$.

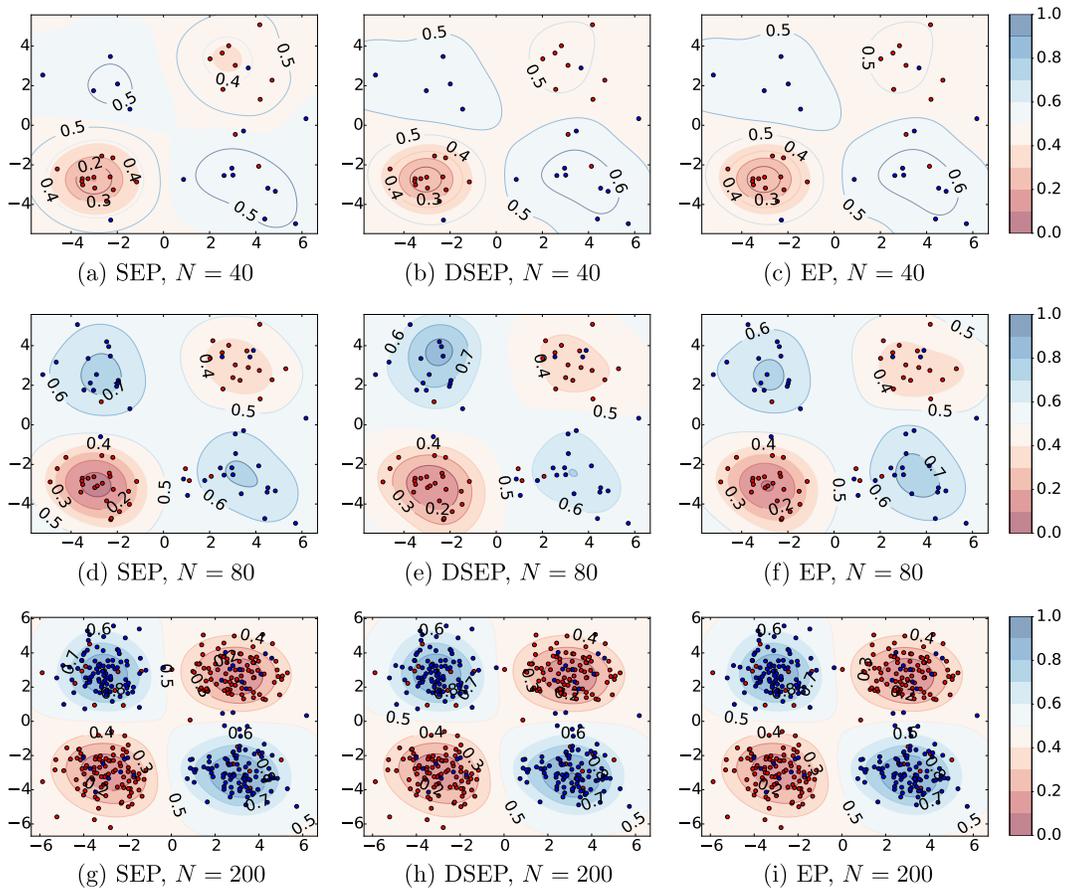

Figure 5: Comparing predictions of kernel probit regression trained by SEP/DSEP/EP, with 10% labels flipped.



\end{document}